\documentclass[11pt]{article}
\usepackage{graphicx}
\usepackage{longtable,lscape}
\usepackage{amsfonts}
\usepackage{float}
\usepackage{url}

\newcommand{\captionfonts}{\footnotesize}
\makeatletter  
\long\def\@makecaption#1#2{%
  \vskip\abovecaptionskip
  \sbox\@tempboxa{{\captionfonts #1: #2}}%
  \ifdim \wd\@tempboxa >\hsize
    {\captionfonts #1: #2\par}
  \else
    \hbox to\hsize{\hfil\box\@tempboxa\hfil}%
  \fi
  \vskip\belowcaptionskip}
\makeatother   
\begin{document}
\title{Measuring Meaning on the World-Wide Web}
\author{Diederik Aerts\\
        \normalsize\itshape
        Center Leo Apostel for Interdisciplinary Studies \\
        \normalsize\itshape
        and Departments of Mathematics and Psychology \\
        \normalsize\itshape
        Vrije Universiteit Brussel, 1160 Brussels, 
       Belgium \\
        \normalsize
        Email: \textsf{diraerts@vub.ac.be}
        }
\date{}
\maketitle              
\begin{abstract}
\noindent
We introduce the notion of the `meaning bound' of a word with respect to another word by making use of the World-Wide Web as a conceptual environment for meaning. The meaning of a word with respect to another word is established by multiplying the product of the number of webpages containing both words by the total number of webpages of the World-Wide Web, and dividing the result by the product of the number of webpages for each of the single words. We calculate the meaning bounds for several words and analyze different aspects of these by looking at specific examples.
\end{abstract}
In previous works, as an aspect of the elaboration of an interpretation of quantum mechanics, we used the World-Wide Web, more specifically, the numbers of webpages containing certain concepts, to show the presence of non-classical, i.e. quantum-like, structures in cognition (Aerts 2009a,b,2010a,b). In the following, we want to apply the experience gained with this use of the World-Wide Web to introduce a means to measure what we will call the `meaning bounds of words with respect to words'.

Let us consider the words `bird' and `feather'. On May 27, 2010, a Yahoo search for the word `bird' returned 705,008,161 hits and a Yahoo search for the words `bird' and `feather', in this order, produced 44,600,421 hits { \url{http://search.yahoo.com/}. This enables us to introduce a quantity which we will call the relative weight of `feather' with respect to `bird' by dividing the above two numbers 42,803,324/705,008,161=0.060713231. In general terms, this means that if we consider a word $A$ and a word $B$, and $n(A)$ is the number of hits for word $A$, while $n(A,B)$ is the number of hits for the words $A$ and $B$, then the relative weight $w(A,B)$ of word $B$ with respect to word $A$ is given by
\begin{equation}
w(A,B)={n(A,B) \over n(A)}
\end{equation} 
Estimations of the total number of webpages of the World-Wide Web vary, because the outcome depends on such factors as the manner of counting and the types of pages considered. For example, when we entered the word `and' into the Yahoo search engine, we found 34,900,551,048 hits. When we entered the word `the', it returned 36,500,597,104 hits, while the number `1' was found 49,001,061,105 times. Most probably, the latter search yielded the largest number of hits we were able to get in this way, since a search for `1' also returns webpages in languages different from English, unlike searches for `the' and `and'. The reason why we insist on using a total number of webpages is that this allows us to introduce a quantity with respect to any word $A$. We have called this quantity the `absolute weight' of this word $A$, i.e. the number of hits divided by the total number $n(\mathrm{www})$ of webpages indexed by the search engine used.
\begin{equation}
w(\mathrm{www},A)={n(A) \over n(\mathrm{www})}
\end{equation}
For the purpose of the present article, after consulting the analysis presented in de Kunder (2010), we decided to set the total number of World-Wide Web pages at $n(www)=55,000,000,000$. The number of hits for the word `feather' is 108,000,638, which means that its absolute weight is given by 108,000,638/55,000,000,000=0.001963648.

The relative weight of `feather' with respect to `bird' is a measure for the relative frequency of appearance of `feather' in the neighborhood of the word `bird', while the absolute weight of `feather' is a measure for the overall relative appearance of `feather' on the World-Wide Web. Hence, if $w(\mathrm{bird,feather})$ is bigger than $w(\mathrm{www,feather})$, this means that `feather' appears more often in the neighborhood of `bird' than it appears generally on the overall World-Wide Web. The fact that `feather' and `bird' are interconnected by a `meaning bound' which is stronger than the meaning bound which connects two arbitrary words implies that $w(\mathrm{bird,feather})$ is bigger than $w(\mathrm{www,feather})$. Indeed, if `bird' and `feather' are connected by a specific meaning bound -- as we know they are --, this connection will statistically give rise to `bird' and `feather' appearing more often in each other's neighborhood than is the case for two arbitrary words. This also means that, if we divide the relative weight of `feather' with respect to `bird' by the absolute weight of `feather', we obtain a quantity which expresses an important aspect of the degree of this meaning bound. For the case of `bird' and `feather', the outcome of this division 30.91859256.

Let us now formulate the equation that governs this important aspect of the meaning bound of words with respect to words. Suppose we consider words $A$ and $B$, and $n(A)$, $n(B)$, $n(A,B)$ are the number of hits for words $A$, $B$ and word $A$ and $B$, respectively. A `meaning bound' $M(A,B)$ of word $B$ with respect to word $A$ is then given by
\begin{equation}
M(A,B)={w(A,B) \over w(B)}={n(A,B)/n(A) \over n(B)/n(www)}={n(A,B)n(www) \over n(A)n(B)}
\end{equation}
Let us consider the words `car' and `world' and calculate their meaning bound as put forward above. We find 4,880,064,558 hits for `car', 11,500,201,838 hits for `world' and 2,234,149,073 hits for `car' and `world'. This gives
\begin{equation}
M(\mathrm{car,world})={2,234,149,073 \cdot 55,000,000,000 \over 4,880,064,558 \cdot 11,500,201,838}=2.189494243
\end{equation}
which is a number slightly bigger than 2. This is much smaller than the number near to 30 that we found for the meaning bound of `feather' with respect to `bird'. And indeed, `bird' and `feather' are much more strongly connected in meaning than `car' and `world' are.

It is not only the meaning bound between single words that our equation allows to calculate. Indeed, we can also consider combinations of words. For example, let us consider the two words `flying' and `air'. The number of hits when the two words are searched for by Yahoo is 376,004,853. The number of hits for the word `bird', as we mentioned already, is 705,008,161, and the number of hits for their combination, i.e. the three words `flying', `air' and `bird', is 56,882,564. These numbers make it possible for us to calculate the meaning bound of `bird' with respect to `flying' and `air' as a combination of words. This gives
\begin{equation}
M(\mathrm{flying;air,bird})={56,882,564 \cdot 55,000,000,000 \over 376,004,853 \cdot 705,008,161}=11.80196318
\end{equation}
Again, this number is much bigger than the one we found as the meaning bound of `world' with respect to `car', but less big than the one of `feather' with respect to `bird'.

Let us consider several special cases to understand better what the meaning bound expresses. For example, we can calculate the meaning bound of a word with respect to itself. Consider again the word `feather', with 108,000,638 hits on the Yahoo search engine. The relative weight of `feather' with respect to `feather' is given by the number of webpages containing `feather' of the set of webpages containing `feather', divided by the number of webpages containing `feather', and hence is equal to 1. The absolute weight of `feather' is given by 108,000,638/55,000,000,000 = 0.001963648. This means that the meaning bound of `feather' with respect to `feather' is given by 55,000,000,000/108,000,638 = 509.2562509, which is a very big number. Remark that the meaning bound of a word with respect to itself is bigger whenever there are fewer websites containing this word. If we consider a combination of words -- a sentence, for example -- which is contained only in one webpage, then the meaning bound of this sentence with respect to itself is 55,000,000,000, which is the value we have chosen for the estimated size of the whole World-Wide Web. More specifically, the meaning bound of an individual webpage with respect to itself is in general equal to 55,000,000,000, the size of the World-Wide Web. The meaning bound of a word that appears in many websites with respect to itself is much less strong. For example, $B(\mathrm{world,world})=4.782524757$, which is only slightly bigger than the meaning bound of `world' with respect to `car'  and much smaller than the meaning bound of `feather' with respect to `bird'.

If the meaning bound $M(A,B)$ of word $B$ with respect to word $A$ equals 1, then the relative weight of $B$ with respect to $A$ equals the absolute weight of $B$, which means that the relative appearance of $B$ neither increases nor decreases in the neighborhood of $A$ as compared to its overall relative appearance on the World-Wide Web. This is why we will call meaning bounds `attractive' when they are bigger than 1, and `repulsive' when they are smaller than 1. We have not yet given an example of a 'repulsive meaning bound'. If we limit ourselves to words of one language, there are not many examples of repulsive meaning bounds, but we can easily find examples of repulsive meaning bounds if we consider two words from different languages. For example, consider the French word `voiture' and the English word `bird', producing 211,003,518 and 705,008,161 hits in the Yahoo search engine, respectively. The number of hits for both words is 1,238,600, which leads to a meaning bound of 0.457941283 of `voiture' with respect to `bird', i.e. a repulsive meaning bound. This means that the word `bird' appears less frequent in the neighborhood of the word `voiture' than it appears overall on the World-Wide Web. However, the meaning bound between words pertaining to different languages is not repulsive by definition. One example is $M(\mathrm{voiture,car})=3.100388372$, since we find 211,003,518 and 4,880,064,558 hits for `voiture' and `car', respectively, and 58,045,516 hits for both words `voiture' and `car' together.

The above example of how `repulsive meaning bounds' appear for words pertaining to different languages, is a clear illustration of the fact that the meaning bound we have introduced measures the degree of `meaning between words' and only indirectly the degree of meaning between the concepts corresponding to these words. This is not unexpected, since the meaning bound is defined by means of `word counts' and not by means of `concept counts'. It remains one of the problems to be solved how to penetrate into the conceptual nature of words, sentences and pieces of text, rather than into the linguistic content defined by the words and their combinations themselves. We will look in more detail into this question when we analyze the meaning bound introduced in the present article with respect to a quantum-mechanical model for the World-Wide Web (Aerts 2010b).

In Table 1 we present the calculations of meaning bounds between several words and combinations of words.
\begin{table}[h]
\scriptsize
\begin{center}
\begin{tabular}{|lllllllllllll|}
\hline 
\multicolumn{1}{|c}{} & \multicolumn{1}{c}{bird} & \multicolumn{1}{c}{car} & \multicolumn{1}{c}{world} & \multicolumn{1}{c}{hierarchy} & \multicolumn{1}{c}{feather} & \multicolumn{1}{c}{chassis} & \multicolumn{1}{c}{bumper} & \multicolumn{1}{c}{math} & \multicolumn{1}{c}{flying;air} & \multicolumn{1}{c}{differential} & \multicolumn{1}{c}{voiture} & \multicolumn{1}{c|}{boomhut} \\
\hline
bird & 78.01 & 3.60 & 2.62 & 5.13 & 30.92 & 1.98 & 5.66 & 7.91 & 10.83 & 3.45  & 0.46 & 0.79 \\
car & 4.20 & 11.27 & 2.19 & 1.86 & 4.03 & 4.73 & 6.63 & 4.48 & 4.48 & 2.91 & 3.15 & 0.57 \\
world & 3.21& 2.35 & 4.78 & 2.54 & 3.63 & 1.16 & 1.68 & 3.36 & 2.98 & 1.54 & 0.38 & 0.67 \\
hierarchy & 4.99 & 1.77 & 2.39 & 686.64 & 6.38 & 1.86 & 2.27 & 10.88 & 6.26  & 16.59  & 0.34 & 0.02 \\
feather & 20.83 & 3.51 & 2.72 & 7.23 & 509.26 & 2.23 & 7.44 & 6.80 & 20.85 & 2.32 & 0.33 & 0.17 \\
chassis & 1.98 & 5.04 & 1.16 & 1.85 & 2.10 & 343.75 & 22.59 & 1.47 & 3.81 & 21.02  & 7.59 & 0.08 \\
bumper & 5.62 & 4.92 & 1.64 & 2.45 & 7.50 & 22.76 & 316.09 & 3.61 & 8.25 & 19.95 & 0.54  & 0.31 \\
math & 8.10 & 3.64 & 2.48 & 11.20 & 19.61 & 1.48 & 4.07 & 133.17 & 7.18 & 21.01  & 1.27 & 0.10 \\
flying;air & 11.80 & 4.23 & 0.01 & 6.58 & 21.70 & 3.94 & 8.28 & 7.56 & 146.27 & 5.07 & 0.91 & 0.16 \\ 
differential & 2.25 & 2.78 & 1.51 & 17.48 & 3.67 & 21.01 & 20.80 & 19.95 & 5.24 & 462.18 & 0.17 & 0.02 \\ 
voiture & 0.46 & 3.10 & 0.39 & 0.34 & 0.33 & 7.54 & 0.54 & 1.21 & 0.91 &  0.17 & 260.66 & 0.06 \\
boomhut & 0.81 & 0.57 & 0.65 & 0.02 & 0.18 & 0.08 & 0.31 & 0.10 & 0.16 &  0.02 & 0.06 & 180916.29 \\ 
\hline
\end{tabular}
\end{center}
\caption{A collection of meaning bound measures for different words}
\end{table}
\normalsize
\noindent
Before we comment on these data, we should briefly point out a specific problem concerning the numbers of webpages yielded by the Yahoo search engine -- as well as by the Google search engine, for that matter. For example, when calculating the meaning bound of `world' with respect to `bird', we found the number of webpages containing `bird' and `world' to be 477,006,321. The number of webpages containing `bird' and `not' containing `world' was 394,003,976. However, the number of webpages containing `bird' was 705,008,161, i.e. much lower than the sum of the number of webpages containing `bird' and `world' (477,006,321) and the number of webpages containing `bird' and `not' containing `world' (394,003,976). So the counts by Yahoo -- and equally so by Google -- were incorrect here. We noticed that this error occurred when making combinations with words that are very abundant on the World-Wide Web, such as `world'. Indeed, the number of webpages containing `world' was 11,500,201,838. To introduce a correction for this error, we proceeded as follows. We divided the number of webpages containing `bird', i.e. 705,008,161, by the sum of the number of webpages containing `bird' and `world' and the number of webpages containing `bird' and `not' containing `world', i.e. 477,006,321 + 394,003,976, which gave a correction factor 0.674640596. We then multiplied this correction factor by the number of webpages containing `bird' and `word' found using Yahoo, which gave us 386,095,722. We consider this `corrected number' to be a fair estimate of the number of webpages containing both words `bird' and `world'. Hence, the meaning bounds of words with respect to other words found in Table 1 have all been established taking into account the above correction.

Overall, the highest values of the meaning bounds are found for words with respect to themselves, but this should not come as a surprise. Obviously, any word will have a strong connection to itself semantically. Let us take a closer look at the word `hierarchy', which we can safely say is not a very usual word. We can see that `hierarchy' has a low meaning bound with respect to words such as `car', `chassis' and `bumper', all of which are common words. It has a high meaning bound with respect to `mathematics' and also with respect to `differential', which are much less common words. And, as those who are familiar with this context will know,, `hierarchy' is indeed a term that makes sense in mathematics, while `differential' is a mathematical term too. Let us consider the word `chassis'. It has a high meaning bound with respect to `car' and `bumper', and indeed both `chassis' and `bumper' are parts of a `car'. It has also a high meaning bound with respect to `differential', and this is interesting to note. Although `differential' is a term from mathematics, it is also a technical part of a `car'. The word `chassis' has also a high meaning bound with respect to the French word `voiture'. And indeed, chassis is also a French word -- although correctly written as follows `ch\^assis', but search engines do not distinguish efficiently between the English and the French way this word is written. The word `mathematics' has a high meaning bound with respect to `hierarchy' and also with respect to `differential', and as we already noted both are indeed mathematical terms. It has also a high meaning bound with respect to `feather' and a relatively high with respect to `bird', which is kind of less obvious -- we could learn eventually why this is the case by looking at a number of the webpages containing both words. The pair of words `flying; air' has a high meaning bound with respect to `bird' and `feather', and it is easy to understand why this is the case, and also a relatively high meaning bound with respect to `bumper' and with respect to `mathematics', which is again less obvious, but would be interesting to find out by looking at a number of the webpages containing the pair of words `flying; air' together with `bumper' or `mathematics'. The word `differential' has a high meaning bound with respect to `hierarchy' and with respect to `mathematics', and also with respect to `chassis' and with respect to `bumper', and we already explained that it is both a term from mathematics and a technical part of a car. The word `voiture' has a high meaning bound with respect to `chassis', and we already explained why this is the case, but also with respect to `car', showing that in this case, if words and their direct translations into another language are considered, the meaning bound measure still applies. The word `boomhut' is the Dutch for `tree house', and like its English counterpart, it is not a very common word. It has a low meaning bound with all words considered, because these words belong to two different languages, Dutch and English -- and Dutch and French for the case of `boomhut' and `voiture'. Interestingly, the highest value of the meaning bound measure between `boomhut' and the other words is reached for the word `bird'. But then there is an obvious relation between `birds' and trees.

We already remarked that the meaning bound of a website with respect to itself equals 55,000,000,000, which is the size of the World-Wide Web. The meaning bound of a website $B$ with respect to another website $A$ equals 0 in general. At least, this is the case if there are no websites containing all the words of website $A$ as well as all the words of website $B$. In the current state of development of the World-Wide Web, this is most probably true. Indeed, in this situation, the relative weight of website $B$ with respect to website $A$ is 0. This is, at first sight at least, not an obvious result. Rather than analyze all the aspects of this state of affairs in the present article, we will shed more light on it within the framework of the quantum-mechanical model for the World-Wide Web we worked out in Aerts (2010b). However, we can here advance two aspects to clarify this state of affairs. First of all, it is related to the as yet relatively small number of webpages that make up today's World-Wide Web. And it is related to the fact that we are concentrating on words rather than on concepts to measure meaning, treating webpages as collections of words, and not as combinations of concepts. Indeed, the human mind, for example, works with schemes of concepts and not with webpages, and quite obviously for any two schemes of concepts there is always at least a third scheme of concepts, and in general very many schemes of concepts, such that both schemes are contained in the third. For such `schemes of concepts', rather than `webpages', the meaning bound between two different schemes of concepts would not be 0 in general, since the schemes are always parts of many other schemes.

We apply the notion of meaning bound of a word with respect to another word, and the analysis of it presented in this article, to identify and investigate the Pet-Fish problem of concept theory (Osherson and Smith 1981), and the conjuntion fallacy in decision theory (Tversky and Kahneman 1983) on the World-Wide Web (Aerts 2010c; Aerts, Czachor, D'Hooghe and Sozzo 2010).

\section*{References}

Aerts, D. (2009a). Quantum particles as conceptual entities. A possible explanatory framework for quantum theory. {\it Foundations of Science}, {\bf 14}, pp. 361-411. 

\smallskip
\noindent
Aerts, D. (2009b). Quantum structure in cognition. {\it Journal of Mathematical Psychology}, {\bf 53}, pp. 314-348.

\smallskip
\noindent
Aerts, D. (2010a). Interpreting quantum particles as conceptual entities. {\it International Journal of Theoretical Physics}.

\smallskip
\noindent
Aerts, D. (2010b). A quantum mechanical model for the World-Wide Web. Preprint Center Leo Apostel, Brussels Free University.

\smallskip
\noindent
Aerts, D. (2010c). The Conjunction Fallacy on the World- Wide Web. Preprint Center Leo Apostel, Brussels Free University.

\smallskip
\noindent
Aerts, D., Czachor, M., D'Hooghe, B. and Sozzo, S. (2010). The Pet-Fish problem on the World-Wide Web. Preprint Center Leo Apostel, Brussels Free University.

\smallskip
\noindent
de Kunder, M. (2010). The size of the World Wide Web. \url{http://www.worldwidewebsize.com/}

\smallskip
\noindent
Osherson, D. N. and Smith, E. E. 1981. On the Adequacy of Prototype Theory as a Theory of Concepts. {\it Cognition}, {\bf 9}, pp. 35-58.

\smallskip
\noindent
Tversky, A. and Kahneman, D. (1983). Extension versus intuitive reasoning: The conjunction fallacy in probability judgment. {\it Psychological Review{, {\bf 90}, pp. 293-315.

\end{document}